\RequirePackage{fix-cm}
\documentclass[smallextended]{svjour3}       

\usepackage{amssymb}
\usepackage{latexsym}
\usepackage{amsmath}
\DeclareMathOperator*{\argmax}{argmax}

\usepackage{natbib}
\usepackage{url}
\usepackage{xcolor}
\definecolor{newcolor}{rgb}{.8,.349,.1}
\usepackage{lineno,hyperref}
\usepackage{lscape}
\usepackage{amsfonts}
\usepackage{gensymb}
\usepackage{subfigure}
\usepackage{array,tabularx}
\usepackage{bm}
\usepackage[Symbol]{upgreek}
\usepackage{xcolor}
\usepackage{amsmath}
\DeclareMathOperator{\sign}{sign}
\usepackage{graphicx}
\usepackage{adjustbox}
\usepackage{caption}
\usepackage{lipsum}
\usepackage[figurename=Figure.,
                  justification=RaggedRight, 
                  labelfont={bf, footnotesize}, 
                  textfont={footnotesize},position=top]{caption}

\begin{document}

\title{Neural Generalization of Multiple Kernel Learning}

\author{Ahmad Navid Ghanizadeh \and
        Kamaledin Ghiasi-Shirazi$^*$\thanks{$^*$ Corresponding author}  \and
        Reza Monsefi \and
        Mohammadreza Qaraei
}

\authorrunning{Ahmad Navid Ghanizadeh et al.} 

\institute{A.N. Ghanizadeh \at
              Department of Computer Science, Saarland University, Saarbrücken, Germany \\
              \email{ahgh00002@uni-saarland.de} 
              \and
          K. Shirazi \at
               Department of Computer Engineering, Ferdowsi University of Mashhad, Mashhad, Iran \\
              \email{k.ghiasi@um.ac.ir}
              \and
              R. Monsefi \at
               Department of Computer Engineering, Ferdowsi University of Mashhad, Mashhad, Iran \\
              \email{monsefi@um.ac.ir}
             \and
              M. Qaraei \at
              Department of Computer Science, Aalto University, Helsinki, Finland \\
              \email{mohammadreza.mohammadniaqaraei@aalto.fi}
}

\date{Received: DD Month YEAR / Accepted: DD Month YEAR}

\maketitle

\begin{abstract}

Multiple Kernel Learning (MKL) is a conventional way to learn the kernel function in kernel-based methods. MKL algorithms enhance the performance of kernel methods. However, these methods have a lower complexity compared to deep models and are inferior to them regarding recognition accuracy. Deep learning models can learn complex functions by applying nonlinear transformations to data through several layers. In this paper, we show that a typical MKL algorithm can be interpreted as a one-layer neural network with linear activation functions. By this interpretation, we propose a Neural Generalization of Multiple Kernel Learning (NGMKL), which extends the conventional MKL framework to a multi-layer neural network with nonlinear activation functions. Our experiments show that the proposed method, which has a higher complexity than traditional MKL methods, leads to higher recognition accuracy on several benchmarks.

    
\end{abstract}


\section{Introduction}
\label{sec:intro}

Kernel methods are a class of machine learning algorithms, which have been widely used in different types of problems \citep{grauman2005pyramid,vedaldi2009multiple,miwa2009protein,longworth2009combining}.
The performance of kernel methods is highly dependent on the type of kernel and the parameters of that kernel. This raises the challenge of learning the kernel function. 
Multiple Kernel Learning (MKL), which learns the combination of a finite set of kernels, is a principled way to address this challenge \citep{lanckriet2004learning,bach2004multiple,sonnenburg2006large,bucak2013multiple,gonen2011multiple}.
In MKL algorithms, a conventional way for kernel learning is to combine the kernel functions linearly.
More sophisticated ways for learning the kernel functions are to use nonlinear combinations \citep{varma2009more,cortes2009learning,xia2012mkboost,zhuang2011two}.
However, it is not clear that these methods can obtain better performance compared to a simple linear combination of kernels, and there is still room for improvement \citep{bucak2013multiple}.  

Similar to other kernel-based methods, two well-known properties of typical MKL algorithms are (i) these methods usually converge to a global solution since they can be formulated as a convex optimization problem, and (ii) kernel functions can be used in large-margin classifiers, for example, Support Vector Machines (SVMs), which reduces overfitting and may lead to better generalization. Although these features have been very appealing in the traditional view of machine learning, their importance has been challenged by the remarkable successes of deep learning methods. 

Deep learning models pass data through several layers, which produces a rich representation of the data and provides a way to learn complex functions. There are remarkable differences between the framework of deep models and kernel methods. In contrast to kernel methods, deep learning models are not usually trained by a convex optimization problem. While it might be considered a drawback, as argued by \citet{bengio2007scaling}, sacrificing convexity may be inevitable for learning complex functions. Furthermore, contrary to kernel methods that use hinge loss, which is margin-based, deep learning models use softmax with cross-entropy in the last layer for classification. However, \citet{rosasco2004loss} have shown that softmax with cross-entropy loss and hinge loss have very similar characteristics, and both entail a loss when the margin is not observed. Therefore, it can be said that deep learning methods that use softmax with cross-entropy are inherently margin-based. Finally, using stochastic gradient descent (SGD) with mini-batches for optimizing deep learning models provides a way for highly parallel computations via GPUs, which has led to state-of-the-art results in large-scale problems.   

In recent years, finding the connection between kernel and deep learning methods has become an active research topic. 
These works have been done mainly with the purpose of boosting kernel methods \citep{cho2010large,mairal2014convolutional,song2018optimizing} or achieving a better understanding of deep learning models \citep{daniely2016toward,jacot2018neural,belkin2018understand}. In this paper, we investigate the connection between deep learning and a set of well-known algorithms in kernel methods, namely MKL, for improving kernel-based algorithms.
We show that the conventional MKL algorithms can be interpreted as a one-layer neural network without nonlinear activation functions in which kernel values represent the input. From this point of view, we propose a Neural Generalization of Multiple Kernel Learning (NGMKL), which extends the shallow linear neural interpretation of MKL to a multilayer neural network with nonlinear activation functions. This practice leads to an MKL model with higher complexity, which improves the capacity of MKL models to learn more complex functions.

MKL models with deep structures have been investigated previously through Multilayer Multiple Kernel Learning (MLMKL) methods \citep{jiu2017nonlinear,jiu2016deep,jiu2019deep}. However, the framework of these methods still sticks to kernelized SVMs, which makes them significantly different from neural networks. More precisely, similar to ordinary MKL methods, MLMKL models use SVMs as classifiers, while neural networks utilize softmax for classification. Secondly, these models have some constraints on the parameters, such as the positivity of the combination weights to keep the deep kernel function positive definite. However, we note that the rationality behind imposing positive definiteness over kernel functions is to obtain a convex objective function, which MLMKL will not attain because of its deep structure. As a result, it may not be beneficial to constrain the kernel function to be a Mercer kernel in a deep structure. Finally, in contrast to typical deep learning models, in MLMKL models, the weights are not usually trained simultaneously, and the training algorithm alternates between learning the kernel combination weights and the classifier's parameters.

In contrast to MKL and MLMKL algorithms, our proposed method utilizes a softmax classifier in the last layer. There are no constraints, such as the positivity of the combination weights, on the parameters of NGMKL, and all the parameters are trained simultaneously. A consequence of the neural interpretation of MKLs is that the framework of deep learning algorithms can be leveraged, which, for instance, provides fast parallel computations via GPUs. Moreover, in the context of SGD, mini-batching allows high throughput since it can process a large number of input examples in many cores at once in GPU. We show that NGMKL outperforms conventional MKL algorithms on commonplace datasets in kernel methods.

The rest of the paper is organized as follows. Section~\ref{sec:related-work} reviews the related work on connecting kernel methods and deep learning. We also review the MLMKL models. In Section~\ref{sec:mkl}, we describe the conventional formulation of MKL algorithms. In Section~\ref{sec:propsed-model}, we show how MKL can be seen in the neural network framework and propose the NGMKL model. The comparison of NGMKL and ordinary MKL algorithms by experiments on several datasets are presented in Section~\ref{sec:experiments}. Finally, we conclude the paper in Section~\ref{sec:conclusion}.

\section{Related Work}
\label{sec:related-work}

Several works in the literature investigate the connections between kernel methods and neural networks. In this section, we review three types of these works: (i) Kernel functions that provably imitate the architecture of multilayer neural networks, (ii) Multilayer neural networks constructed by kernel approximation techniques, and (iii) MKL methods with multilayer structures.

Some methods link kernel methods with neural networks by proposing kernel functions that mimic the computation of neural networks. For example, similar to neural networks, it has been demonstrated that the feature map of arc-cosine kernels consists of a linear transformation using an infinite number of weights followed by a nonlinearity \cite{cho2010large}. However, the interpreted weights are fixed and have a Gaussian distribution with zero mean and unit variance. 
To tackle this drawback, \citet{pandey2014learning} proposed to learn the covariance of the interpreted weights by stretching a finite number of trained weights to an infinite number.
\cite{pandey2014go} proposed learning the covariance by an iterative approach similar to restricted Boltzmann machines. In arc-cosine kernels, one can change the nonlinearity of the interpreted neural network and extend the structure to multiple layers by using two distinct hyperparameters. The connections between neural networks with an infinite number of weights and kernel methods are also discussed in the context of Gaussian processes, which has attracted attention for a better understanding of deep learning in recent years \citep{hazan2015steps,wilson2016deep,lee2017deep,jacot2018neural}.

While kernel functions are implicitly seen as an infinite-width neural network in the aforementioned methods, some methods use kernel approximation techniques \citep{williams2001using,rahimi2008random} to explicitly construct one layer of a neural network by a kernel function. For example, convolutional kernel networks that link convolutional neural networks with kernel methods are constructed by consecutively approximating a convolution kernel \citep{mairal2014convolutional,mohammadnia2018convolutional,mairal2016end}. In another method called M-DKMO, a neural network is used upon the feature maps of some kernel functions approximated by the Nystr\"{o}m method for kernel learning in an end-to-end fashion \citep{song2018optimizing}. Although this method uses neural networks for learning kernels, it differs from the proposed NGMKL method. NGMKL reformulates the MKL problem through neural networks, while M-DKMO uses neural networks for learning approximated kernel feature maps.

More similar to our work, some methods proposed MKL problems in which the kernels are combined through a neural-network-like architecture. These methods, which are known as Multilayer Multiple Kernel Learning (MLMKL) models, were first introduced by \cite{zhuang2011two}. 
In the method of \citet{zhuang2011two}, the kernel function is defined as a convex combination of a set of base kernels followed by an exponential nonlinearity, which can be seen as a two-layer MKL problem. \citet{rebai2016deep} extended two-layer MLMKLs to several layers and optimized the combination parameters by backpropagation. \citet{jiu2017nonlinear} utilized unsupervised, semi-supervised, and Laplacian SVM as the objective functions for MLMKL. For reducing computations and improving the scalability of MLMKL, \citet{jiu2019deep} proposed kernel approximation in a greedy layer-wise fashion. To make MLMKL more similar to neural networks, \citet{sahbi2019totally} formulated the base kernels as a neural operation and optimized the support vectors of MLMKLs along with the other parameters.

Although MLMKL models combine multiple kernels in a multilayer structure, NGMKL differs since the multilayer neural architecture in MLMKL is only for combining the base kernels. In other words, the kernel combination in MLMKL has a multilayer structure, while in our proposed method, the whole MKL problem, namely kernel combination and classification, is interpreted as a single neural network that can be extended to a deep structure. As a result, firstly, MLMKLs use SVMs for classification, while NGMKLs utilize softmax with cross-entropy, similar to ordinary neural networks. Secondly, MLMKL has some constraints on the parameters, whereas NGMKL is formulated as an unconstrained optimization problem. Finally, in most of the MLMKL models, the parameters for combining the kernels and the parameters of the classifier are learned alternatively. In contrast, in NGMKL, all the parameters are learned simultaneously.

\section{Multiple Kernel Learning}
\label{sec:mkl}

In a typical kernel-based method, data is represented by a positive definite function called a kernel, which computes the similarity between data. However, a single kernel might not be rich enough to capture the complex underlying patterns of real-world data. In MKL algorithms, data is represented by several kernels, and the aim is to learn the optimal combination of these representations. Using multiple kernels instead of a single kernel increases the learning capacity of kernel-based models. In this section, we describe the conventional framework of MKL models.

Let $D=\{(\textrm{x}_1,y_1),...,(\textrm{x}_n,y_n)\}$ consist of $n$ pairs of feature vectors $\textrm{x}_i \in \mathbb{R}^{d}$ and their corresponding labels $y_i \in \{-1,+1\}$. Using a kernel $k(.,.)\rightarrow R$, each input $\textrm{x}$ is represented by the vector $\textrm{k}^{\textrm{x}}=[k(\textrm{x},\textrm{x}_1),...,k(\textrm{x},\textrm{x}_n)]^T$. Then a linear function $f:\mathbb{R}^{n}\rightarrow\mathbb{R}$ parameterized by $\gamma \in \mathbb{R}^{n}$ could be used on top of $\textrm{k}^\textrm{x}$ to predict the label of $\textrm{x}$:
\begin{equation}
\label{eq:map-K}
f(\textrm{k}^\textrm{x}) = \gamma^T \textrm{k}^\textrm{x}
\end{equation}

By using $s$ different base kernels $k_1,...,k_s$, the data can be represented by vectors $\textrm{k}^\textrm{x}_{1},...,\textrm{k}^\textrm{x}_{s}$ corresponding to these kernels. 
If these vectors are passed through the function $f$, the final classifier $f_c:\mathbb{R}^{n}\rightarrow\{-1,1\}$ is formed by combining the output of these functions:
\begin{equation}
\label{eq:MKL-classify}
f_c(\textrm{x}) = \sign(\sum_{j=1}^s \beta_j f(\textrm{k}^\textrm{x} _j)),
\end{equation}
where $\beta_j$ can be interpreted as the weight of the kernel $k_j$.
In support vector machines, which seek to find a hyperplane with the maximum margin, the MKL problem can be formulated as the following optimization problem \citep{sonnenburg2006large}:
\begin{equation}
\label{eq:minmax}
\begin{aligned}
&\max_\beta \min_\alpha
& & {1\over2} \sum_{j=1}^s \beta_j (\alpha\odot \textrm{y})^T G_j(\alpha\odot \textrm{y})-\alpha^T\mathbf{1} \\
& \text{ s.t.} & &  \sum_{i=1}^n \alpha_i y_i =0, \\
& & &  \sum_{j=1}^S \beta_j = 0, \\
& & &  \displaystyle 0 \leq \alpha  \leq C, \\
& & &  \beta_j \geq 0 ,\quad \forall j=1,...,s \\
\end{aligned}
\end{equation}
where $G_j=[\textrm{k}^{\textrm{x}_1}_j,...,\textrm{k}^{\textrm{x}_n}_j]$, $\textrm{y}=[y_1,...,y_n]^T$, $\alpha \in \mathbb{R}^{n}$, $\mathbf{1}$ is a vector of ones, and $C$ is the hyperparameter of the SVM. The decision function obtained by solving Eq.~(\ref{eq:minmax}) is similar to Eq.~(\ref{eq:MKL-classify}) in which $\gamma=\alpha \odot \textrm{y}$.

\section{Proposed Model}
\label{sec:propsed-model}
In this section, we propose a formulation of the MKL problem as a neural network.  
The core element of a neural network is a basic module in which a linear operation followed by a nonlinear one are applied to the input:
\begin{equation}
\label{eq:module}
    \hat{\textrm{u}} = \phi(\Upgamma^T \textrm{u})
\end{equation}
where $\textrm{u} \in \mathbb{R}^n$ is a feature vector, $\Upgamma \in \mathbb{R}^{n\times n^\prime}$ is a matrix of parameters, $\phi(.)$ is a nonlinear activation function, and $\hat{\textrm{u}}  \in \mathbb{R}^{n^\prime}$ is the output of the module.

In MKL, multiplication of each kernelized representation of data, $\textrm{k}^\textrm{x}$, with the parameter vector $\gamma \in \mathbb{R}^n$ (Eq.~(\ref{eq:map-K})) can be seen as a basic module in a neural network but without nonlinearity. By applying a nonlinear function $\phi(.)$, Eq.~(\ref{eq:map-K}) becomes equivalent to this module in a neural network:
\begin{equation}
\label{eq:nonlinear_nn}
    z = \phi(\gamma^T \textrm{k}^\textrm{x})
\end{equation}

\begin{figure*}[pt!]
    \centering
    \includegraphics[width=14.4cm]{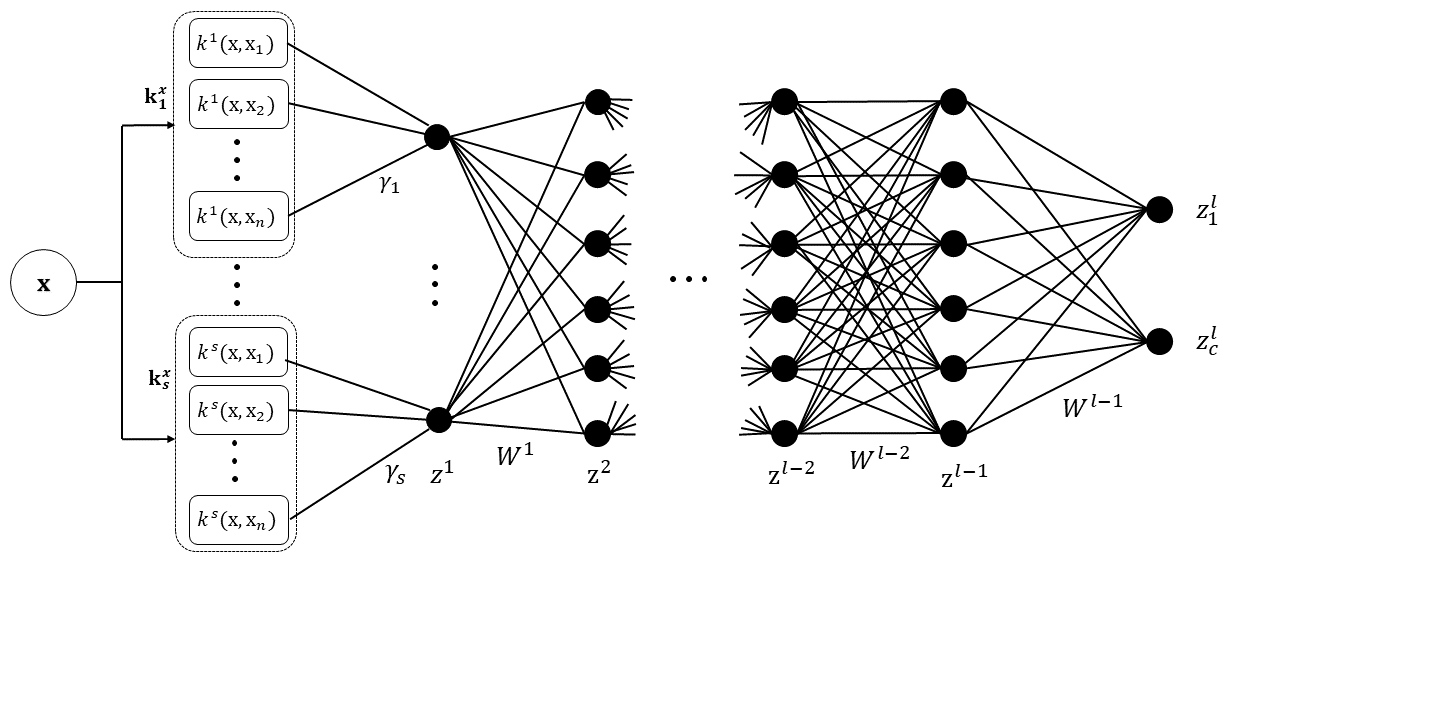}
    \caption{Architecture of the proposed network with $L$ layers.} 
    \label{fig:Architecture}
\end{figure*}

Having multiple kernel representations $\textrm{k}_1^{\textrm{x}},...,\textrm{k}_s^{\textrm{x}}$ for data $\textrm{x}$ obtained by different kernels, by Eq.~(\ref{eq:nonlinear_nn}), we have parameters $\gamma_1,...,\gamma_s$ and representations $z_1^1,...,z_s^1$ corresponding to these kernels. We concatenate $z_1^1,...,z_s^1$ to form $\textrm{z}^1$ as follows:
\begin{equation}
\label{eq:concat}
    \textrm{z}^1=[z_1^1,...,z_s^1]^T
\end{equation}
Then we apply the basic module of neural networks to $\textrm{z}^1$ repeatedly as follows:
\begin{equation}
\label{eq:NGMKL}
    \textrm{z}^{l} = \phi_l((W^{l-1})^T ...\phi_1((W^1)^T \textrm{z}^1)...) ,
\end{equation}
where $W^1 \in \mathbb{R}^{ s\times n_1}$ and $W^{i} \in \mathbb{R}^{n_{i-1}\times n_i}$  for $i>1$ are weight matrices, and $\phi_i(.)$ is an activation function. The above equation generalizes the MKL problem to a deep neural network, which we call the Neural Generalization of Multiple Kernel Learning (NGMKL). In particular, in a binary classification problem, by setting $l=1$ and setting the same parameters for $\gamma_1,..., \gamma_s$, and using a linear activation function as $\phi_1$, the structure of MKL problem of Eq.~(\ref{eq:NGMKL}) becomes equivalent to the conventional MKL framework. The architecture of NGMKL is shown in Figure~\ref{fig:Architecture}.

While conventional MKL frameworks use SVM for classification, in NGMKL, we use softmax for classification. Let $\textrm{z}^{l} \in \mathbb{R}^c$, where $c$ is the number of classes. The softmax classifier assigns $c$ probabilities to the data as follows:
\begin{equation}
\label{eq:softmax}
     \hat{y}_i = \frac{e^{z_i^{l}}}{\sum_{j=1}^c e^{z_j^{l}}},
\end{equation}
and the label of the data is defined as $k = \argmax\limits_i \hat{y}_i$.

For training NGMKL, we use the cross-entropy loss function. Assume $D=\{(\textrm{x}_1,\textrm{y}_1),...,(\textrm{x}_n,\textrm{y}_n)\}$ consists of $n$ pairs of data $\textrm{x}_i \in \mathbb{R}^{d}$ and the corresponding labels $\textrm{y}_i \in \{0,1\}^c$, where $y_{ik}=1$ if $k$ is the relevant class for the sample $\textrm{x}_i$, and $y_{ik}=0$ otherwise. The cross-entropy loss function for a sample $(\textrm{x},\textrm{y})$, in which $y_k=1$ and $y_{i\neq k}=0$, is defined as follows:
\begin{equation}
\label{eq:cross-entropy}
    J = -z_k^{l} + \log \left({\sum_{j=1}^c e^{z_j^{l}}}\right)
\end{equation}

We train all the parameters of NGMKL, including $W^1,...,W^l$ and $\gamma_1,...,\gamma_s$, simultaneously, using the backpropagation algorithm. The derivatives of the cross-entropy function with respect to the parameters of a two-layer NGMKL are as follows:
\begin{equation}
\label{eq:derivativeW}
    \frac{\partial J}{\partial w^1_{ij}} = (\hat{y_j}-y_j)\phi_1(z^1_i)
\end{equation}

\begin{equation}
\label{eq:derivativegamma}
    \frac{\partial J}{\partial \gamma_i} = \textrm{w}^1_{i*}(\textrm{y}-\hat{\textrm{y}})\phi_1 ^ \prime (z_i^1) \textrm{k}_\textrm{x}^i
\end{equation}
where $\textrm{w}^1_{i*}$ is the $i$-th row of $W^1$ and $\hat{\textrm{y}}=[\hat{y_1},...,\hat{y_c}]^T$.

As seen in Eq.~(\ref{sec:mkl}), the SVM framework of MKL has several constraints on the parameters. These constraints are for preserving the convexity of the problem and ensuring a large margin. However, in our proposed model, there is no constraint on the parameters, and the goal of large-margin classification is obtained by a combination of softmax with cross-entropy loss and weight decay (which penalizes large weights). In Section~\ref{sec:experiments}, we show that although NGMKL is formulated as a non-convex optimization problem, it can outperform the typical MKL models.
\begin{table}[t]				
	\caption{Summary of the datasets used in the experiments from the UCI repository.}	
	\label{tab:datasetinfo}			
	\centering
	\footnotesize
	\begin{tabular}{|l|c|c|c|}
		\hline
		 Dataset & \# Training set & \# Test set & Dimensions \\
		\hline
		   a1a & $802$ & $803$ & $123$\\
		\hline
		   a3a & $1592$ & $1593$ & $123$\\
		\hline
		   a4a & $2391$ & $2392$ & $123$\\
		\hline
		   german.numer & $500$ & $500$ & $24$\\
		\hline
		   ionosphere & $175$ & $176$ & $34$\\
		\hline
		   monks1 & $556$ & $557$ & $6$\\
		\hline
		   sonar & $104$ & $104$ & $60$\\
		\hline
		   svmguide3 & $621$ & $622$ & $21$\\
		\hline
		 banana & $400$ & $4900$ & $2$\\
		\hline
		 b.cancer & $200$ & $77$ & $9$\\
		\hline
		 diabetes & $468$ & $300$ & $8$\\
		\hline
		 german & $700$ & $300$ & $20$\\
		\hline
		 image & $1300$ & $1010$ & $18$\\
		\hline
		 ringnorm & $400$ & $7000$ & $20$\\
		\hline
		 f.solar & $666$ & $400$ & $9$\\
		\hline
		 thyroid & $140$ & $75$ & $5$\\
		\hline
		 titanic & $150$ & $2051$ & $3$\\
		\hline
		 twonorm & $400$ & $7000$ & $20$\\
		\hline
		 waveform & $400$ & $4600$ & $21$\\
		\hline
	\end{tabular}
\end{table}
\section{Experimental Results}
\label{sec:experiments}

In this section, we evaluate NGMKL in several classification problems and compare it with conventional MKL models. We use a two-layer NGMKL and three variants of this model, namely "NGMKL1", "NGMKL2", and "NGMKL3", which are different from each other in the choice of the input kernel.
In the following subsections, we describe the datasets we used for the evaluation of the models, the baseline models with which we compare NGMKL, the architecture and the optimization settings of NGMKL, and the three variants of NGMKL which are different in the choice of the input kernel.

\subsection{Datasets}
\label{sec:datasets}
Since the conventional MKL models are not scalable, we use UCI datasets to evaluate the models. 
These datasets are described in Table~\ref{tab:datasetinfo}.
Following \cite{ratsch2001soft}, each dataset is randomly split into the training and testing data, and this practice is repeated 100 times, except for the image dataset, which has 20 splits.

\subsection{Baselines}
\label{sec:baselines}
We compare NGMKL with some well-known MKL algorithms:

\begin{itemize}
    \item MKL-SD: an $\ell_1$-norm regularized MKL solved by subgradient descent methods \citep{rakotomamonjy2008simplemkl}.
    \item MKL-SILP: an $\ell_1$-norm regularized MKL with a semi-infinite linear programming objective function \citep{sonnenburg2006general}.
    \item MKL-Hessian: an MKL algorithm solved by second-order optimization methods \citep{chapelle2008second}. 
    \item $\ell_p$-MKL: an $\ell_p$-norm regularized MKL in which we set $p=2$ \citep{kloft2011lp}. 
    \item MKBoost: A boosting approach to the MKL problem \citep{xia2012mkboost}.
\end{itemize}

\subsection{NGMKL setup}
We use a two-layer NGMKL in all the experiments. The nonlinearity of the hidden layer is the scaled hyperbolic tangent $f(x)=1.7159\tanh({\frac{2}{3}x})$. We use the Stochastic Gradient Descent (SGD) algorithm to train our model in which the batch size is 40, the learning rate is 0.01, and the weight decay is $5\times 0^{-6}$.

\subsection{Three variants of NGMKL}
\label{sec:variants}
We use three variants of NGMKL named ``NGMKL1'', ``NGMKL2'', and ``NGMKL3'' regarding the methods used for selecting the input kernels from the base kernels. There are 17 base kernels in all the models, including NGMKLs and the baselines: three polynomial kernels ($k(\textrm{x},\textrm{y})=(\textrm{x}^T\textrm{y})^d$) of degrees 1, 2, and 3 as well as 14 Gaussian kernels ($k(\textrm{x},\textrm{y})=e^{\frac{-1}{2\sigma^2}\Vert \textrm{x}-\textrm{y}\Vert _2 ^2}$) with kernel widths $\sigma = 2^{-6},...,2^7$. The three variants of NGMKL are as follows: 
\begin{itemize}
    \item NGMKL1: All 17 base kernels are used as the input kernels in this model.
    \item NGMKL2: In this model, $\ell_1$-MKL \citep{sonnenburg2006general} is used for selecting the input kernels from the base kernels. $\ell_1$-MKL produces sparse weights for combining the base kernels by penalizing the $\ell_1$ norm of the weights.
    \item NGMKL3: MKBoost-D1 \citep{xia2012mkboost} is used for selecting the input kernels among the 17 base kernels. MKBoost-D1, in each trial of the boosting method, selects the kernel with the minimum error and adds it to the set of selected kernels. The selection is with replacement.
\end{itemize}
\begin{table}[t!]				
	\caption{The results of our three proposed methods [mean classification error $\pm$ std] in percent on UCI datasets. The best performance in each dataset is highlighted in \textbf{bold}.}	
	\label{tab:Threebgmkl}			
	\centering
	\footnotesize
	\begin{tabular}{|l|c|c|c|}
		\hline
		Dataset & NGMKL1 & NGMKL2 & NGMKL3 \\
		\hline
		 a1a & $18.04\pm 0.00 $ & \bm{$15.94\pm 0.00 $} & $17.39\pm 0.01 $ \\
		\hline
		a3a	& $17.93\pm 0.00 $ & $17.56\pm 0.00 $ & \bm{$17.08\pm 0.01 $} \\
		\hline
		a4a & $18.22\pm 0.00 $ & $19.01\pm 0.00 $ & \bm{$17.32\pm 0.01 $} \\
		\hline
		german.numer & \bm{$24.57\pm 0.01 $}& $25.03\pm 0.01 $  & $25.79\pm 0.02 $ \\
		\hline
		ionospher & $06.99\pm 0.00 $ & $05.63\pm 0.00 $ & \bm{$05.09\pm 0.30 $} \\
		\hline
		monks1 & \bm{$04.87\pm 0.09 $} & $05.16\pm 0.00 $ & $04.91\pm 0.03 $ \\
		\hline
		sonar & $21.41\pm 0.08 $ & $20.78\pm 0.05 $ & \bm{$17.93\pm 0.23 $} \\
		\hline
		svmguide3 & $18.45\pm 0.00 $ & \bm{$17.73\pm 0.00 $} & $18.08\pm 0.02 $ \\
		\hline
		banana & $12.00\pm 0.00 $ & \bm{$11.00\pm 0.33 $} & $11.06\pm 0.00 $ \\
		\hline
		b.cancer  & $26.49\pm 0.30 $ & $27.86\pm 0.17 $ & \bm{$26.04\pm 0.22 $} \\
		\hline
		diabete  & $24.40\pm 0.02$ & $26.40\pm .0.05 $ & \bm{$24.29\pm 0.03 $} \\
		\hline
		german  & $24.19\pm 0.05 $ & \bm{$23.83\pm 0.05 $} & $23.91\pm 0.05 $ \\
		\hline
		image & $03.55\pm 0.00 $ & $04.06\pm 0.00 $ & \bm{$03.14\pm 0.00 $} \\
		\hline
		ringnorm & $02.07\pm 0.00 $ & $01.96\pm 0.00 $ & \bm{$01.57\pm 0.00 $} \\
		\hline
		f.solar & $35.80\pm 0.04 $ & \bm{$35.57\pm 0.07 $} & $35.79\pm 0.03 $ \\
		\hline
		thyroid & $05.53\pm 0.10 $ & $05.80\pm 0.05 $ & \bm{$04.17\pm 0.06 $} \\
		\hline
		titanic  & $22.34\pm 0.00 $ & \bm{$22.12\pm 0.00 $} & $22.41\pm 0.01 $ \\
		\hline
		twonorm  & \bm{$02.80\pm 0.00 $} & $03.05\pm 0.00 $ & $02.90\pm 0.00 $ \\
		\hline
		waveform & \bm{$09.37\pm 0.01 $} & $10.68\pm 0.04 $ & $09.55\pm 0.00 $ \\
		\hline
		
	\end{tabular}				
\end{table}

\subsection{Comparison results}
A comparison of three variants of NGMKLs is shown in Table~\ref{tab:Threebgmkl}. 
This table compares three variants of NGMKLs, showing that all NGMKL methods generally achieve similar results. However, NGMKL3 achieves better results compared to the other variants of NGMKLs in 9 datasets. The second successful variant is NGMKL2, achieving better results in 6 datasets, and NGMKL1 has a better error rate in the remaining datasets.

Table~\ref{tab:resultMKL} compares popular MKL algorithms with NGMKL3. 
Comparing NGMKL3 and the best competing method for each row, we have highlighted statically significant results, using an unpaired $t$ test, in boldface. Results for ``monks'' and ``sonar'' datasets are not statistically significant. Overall for 17 datasets, the differences between NGMKL3 and the best other method are statistically significant.

\begin{landscape}
\begin{table} [ht] 
	\caption{The average testing errors [mean classification error $\pm$ std] in percent of five popular MKL Algorithms and one of the proposed methods (NGMKL3). The relative ranking of each algorithm on each data set is shown in the parenthesis. Rank 1 refers to the algorithm which achieves the highest accuracy on the given data set.}	
	\label{tab:resultMKL}			
	\centering
	\footnotesize
    \begin{adjustbox}{width=1.6\textwidth,center=\textwidth}
	
	\begin{tabular}{|l|c|c|c|c|c|c|}
		\hline
		Dataset & MKL-SD & MKL-SILP & MKL-Hessian & MKboost & $\ell_p$-MKL &  NGMKL3\\
		\hline
		  a1a & $24.53\pm 0.09 (4)$ & $24.53\pm 0.09 (4)$ & $24.56\pm 0.08 (5)$ & $18.78\pm 0.77 (2)$ & $23.93\pm 0.66 (3)$ & $\mathbf{17.39\pm 0.01 (1)}$\\
		\hline
		  a3a & $24.24\pm 0.03 (4)$ & $24.24\pm 0.03 (4)$ & $24.55\pm 0.00 (5)$ & $18.63\pm 0.51 (2)$ & $24.03\pm 0.40 (3)$ & $\mathbf{17.08\pm 0.01 (1)}$\\
		\hline
		 a4a & $24.85\pm 0.01 (4)$ & $24.85\pm 0.01 (4)$ & $24.85\pm 0.01 (4)$ & $18.90\pm 0.52 (2)$ & $24.39\pm 0.33 (3)$ & $\mathbf{17.32\pm 0.01 (1)}$\\
		\hline
		 german.numer & $29.99\pm 0.04 (5)$ & $29.97\pm 0.25 (4)$ & $30.00\pm 0.06 (6)$ & $26.39\pm 1.17 (2)$ & $29.35\pm 0.55 (3)$ & $\mathbf{25.79\pm 0.02 (1)}$\\
		\hline
		  ionospher & $06.94\pm 1.55 (5)$ & $07.51\pm 2.55 (6)$ & $06.71\pm 1.77 (4)$ & $05.74\pm 1.39 (3)$ & $05.63\pm 2.00 (2)$ & $\mathbf{05.09\pm 0.30 (1)}$\\
		\hline
		  monks & $20.90\pm 2.51 (5)$ & $20.68\pm 3.11 (4)$ & $17.41\pm 2.66 (3)$ & $04.78\pm 1.40 (1)$ & $21.56\pm 0.00 (6)$ & $04.91\pm 0.03 (2)$\\
		\hline
		  sonar & $22.48\pm 4.23 (5)$ & $22.09\pm 4.27 (4)$ & $24.13\pm 3.65 (6)$ & $18.17\pm 3.96 (2)$ & $19.76\pm 4.06 (3)$ & $17.93\pm 0.23 (1)$\\
		\hline
		 svmguide3 & $22.56\pm 0.38 (4)$ & $22.58\pm 0.40 (5)$ & $22.83\pm 0.00 (6)$ & $18.42\pm 0.67 (2)$ & $21.84\pm 0.57 (3)$ & $\mathbf{18.08\pm 0.02 (1)}$\\
		\hline
		banana  & $10.71\pm 0.00 (2)$ & $\mathbf{10.38\pm 0.00 (1)}$ & $10.86\pm 0.00 (3)$ & $12.12\pm 0.00 (6)$ & $11.95\pm 0.00 (5)$ & $11.06\pm 0.00 (4)$\\
		\hline
		b.cancer  & $28.38\pm 0.25 (4)$ & $28.31\pm 0.26 (3)$ & $27.53\pm 0.22 (2)$ & $31.45\pm 0.18 (5)$ & $33.51\pm 0.11 (6)$ & $\mathbf{26.04\pm 0.22 (1)}$\\
		\hline
		diabetes  & $24.68\pm 0.07 (3)$ & $24.88\pm 0.06 (4)$ & $\mathbf{23.60\pm 0.02 (1)}$ & $28.65\pm 0.12 (6)$ & $24.93\pm 0.05 (5)$ & $24.29\pm 0.03 (2)$\\
		\hline
		german  & $27.83\pm 0.12 (4)$ & $27.42\pm 0.11 (3)$ & $23.93\pm 0.04 (2)$ & $28.09\pm 0.06 (5)$ & $30.90\pm 0.08 (6)$ & $\mathbf{23.91\pm 0.05 (1)}$\\
		\hline
		image  & $05.34\pm 0.00 (5)$ & $05.34\pm 0.00 (5)$ & $\mathbf{02.05\pm 0.00 (1)}$ & $02.87\pm 0.00 (2)$ & $03.68\pm 0.00 (4)$ & $03.14\pm 0.00 (3)$\\
		\hline
		ringnorm  & $02.11\pm 0.00 (3)$ & $02.13\pm 0.00 (4)$ & $01.71\pm 0.00 (2)$ & $05.16\pm 0.00 (6)$ & $03.037\pm 0.00 (5)$ & $\mathbf{01.57\pm 0.00 (1)}$\\
		\hline
		f.solar  & $32.45\pm 0.02 (2)$ & $32.45\pm 0.02 (2)$ & $35.38\pm 0.02 (3)$ & $36.49\pm 0.03 (5)$ & $\mathbf{32.42\pm 0.03 (1)}$ & $35.79\pm 0.03 (4)$\\
		\hline
		thyroid  & $04.21\pm 0.04 (3)$ & $04.47\pm 0.04 (4)$ & $04.20\pm 0.04 (2)$ & $04.71\pm 0.05 (5)$ & $05.73\pm 0.11 (6)$ & $\mathbf{04.17\pm 0.06 (1)}$\\
		\hline
		titanic  & $22.32\pm 0.00 (4)$ & $22.27\pm 0.00 (3)$ & $22.17\pm 0.00 (2)$ & $22.45\pm 0.03 (6$ & $\mathbf{22.09\pm 0.00 (1)}$ & $22.41\pm 0.01 (5)$\\
		\hline
		twonorm  & $02.99\pm 0.00 (2)$ & $03.02\pm 0.00 (3)$ & $03.36\pm 0.00 (4)$ & $03.56\pm 0.00 (5)$ & $03.90\pm 0.00 (6)$ & $\mathbf{02.90\pm 0.00 (1)}$\\
		\hline
		waveform  & $11.84\pm 0.00 (3)$ & $11.89\pm0.00 (4)$ & $11.00\pm 0.00 (2)$ & $12.13\pm 0.00 (5)$ & $14.15\pm 0.00 (6)$ & $\mathbf{09.55\pm 0.00 (1)}$\\
		\hline
	\end{tabular}
	\end{adjustbox}
\end{table}
\end{landscape}

\begin{figure*}[pt!]
    \centering
    \includegraphics[width=11.7cm]{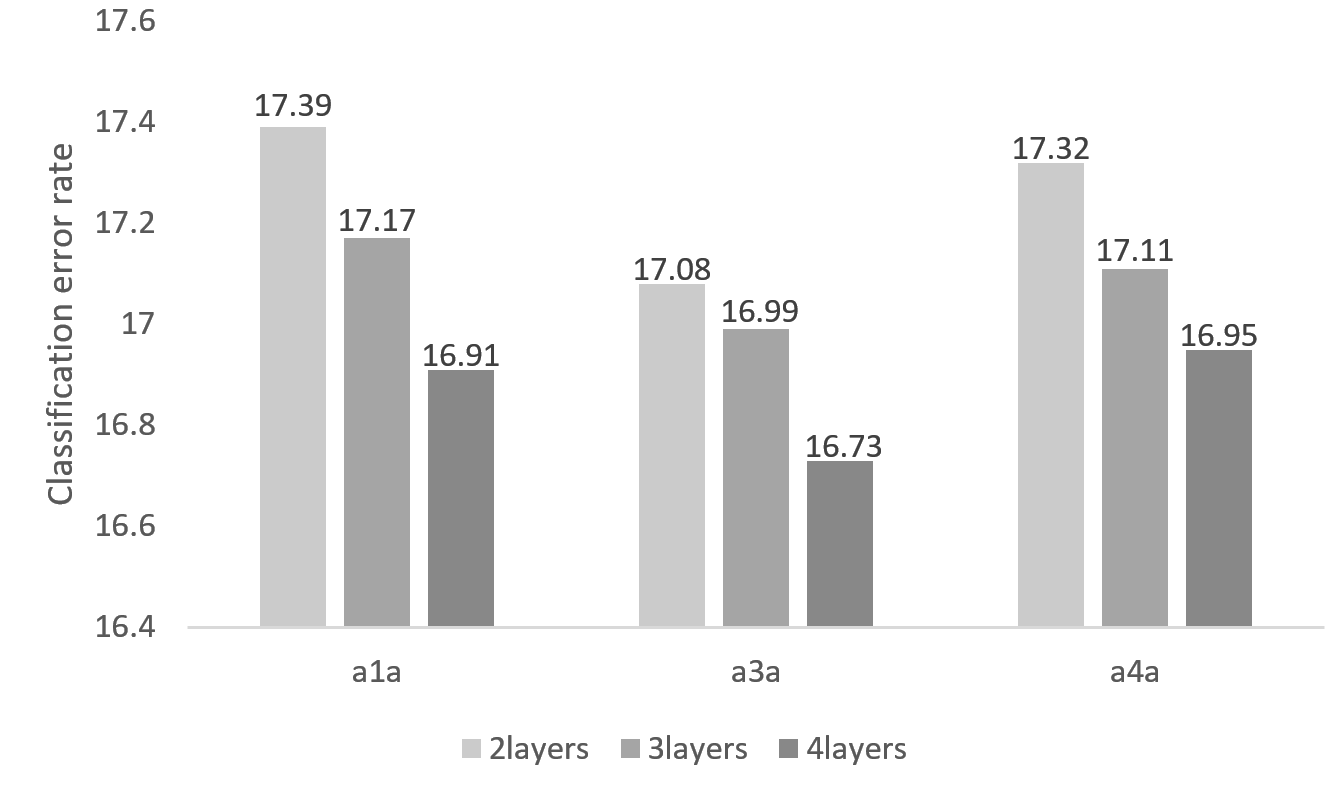}
    \caption{Evaluating the performance of NGMKL3 with different numbers of layers.} 
    \label{fig:diff_layers}
\end{figure*}
According to Table~\ref{tab:resultMKL}, all regular MKL algorithms have the same error rate because they have the same formulation but adopt different strategies to speed up the optimization process. However, NGMKL3 is superior to the traditional MKL methods in most datasets (12 out of 17). 

In the next step, we evaluate NGMKL3  with different numbers of layers. For simplicity, we only compare with three different layers (L=2,3,4). When we have a higher number of layers, most of the datasets remain unchanged in terms of the error rate (16 out of 19), and only three of them (namely ala, a3a, and a4a) have a decrease in the classification error rate (Figure~\ref{fig:diff_layers}). As seen from Table~\ref{tab:datasetinfo}, these three datasets have more dimensions than others. Since our proposed method (NGMKL3) can learn complex functions by extending the layers with nonlinear activation functions, the classification error on these three datasets has decreased. However, extending the layers cannot improve accuracy in most datasets, especially when using UCI datasets, since most of them are not complex enough. 
In addition, the three variants of our proposed methods are not sensitive to the number of layers; therefore, we evaluated our proposed methods in comparison with conventional methods based on two layers.

\begin{figure*}[pt!]
    \centering
    \includegraphics[width=11.9cm]{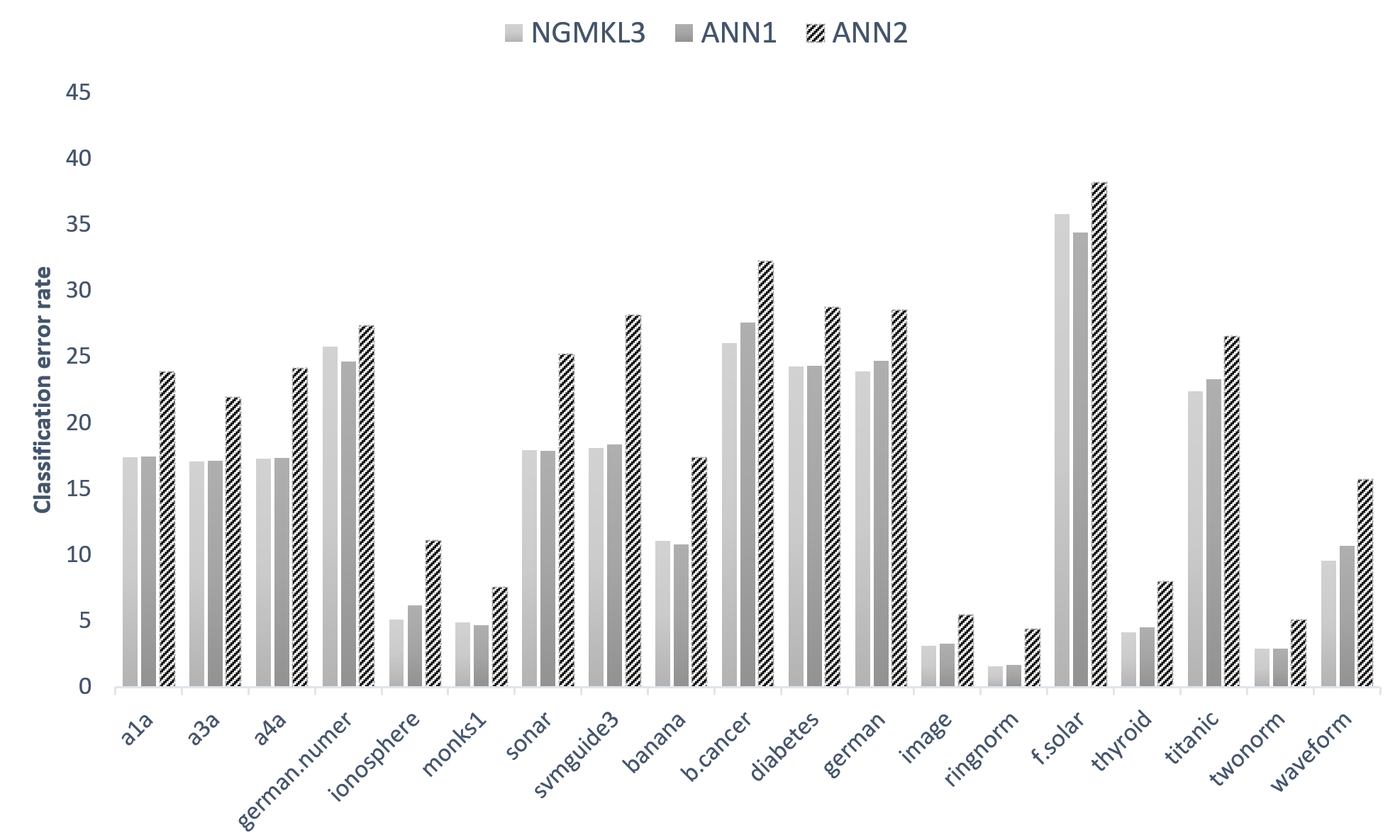}
    \caption{Evaluating the performance of NGMKL3 with two variants of Artificial Neural Network (ANN) with two layers.} 
    \label{fig:NNMKL}
\end{figure*}

\begin{figure*}[pt!]
    \centering
    \begin{minipage}[b]{0.45\textwidth}
         \centering
         \includegraphics[width=\textwidth]{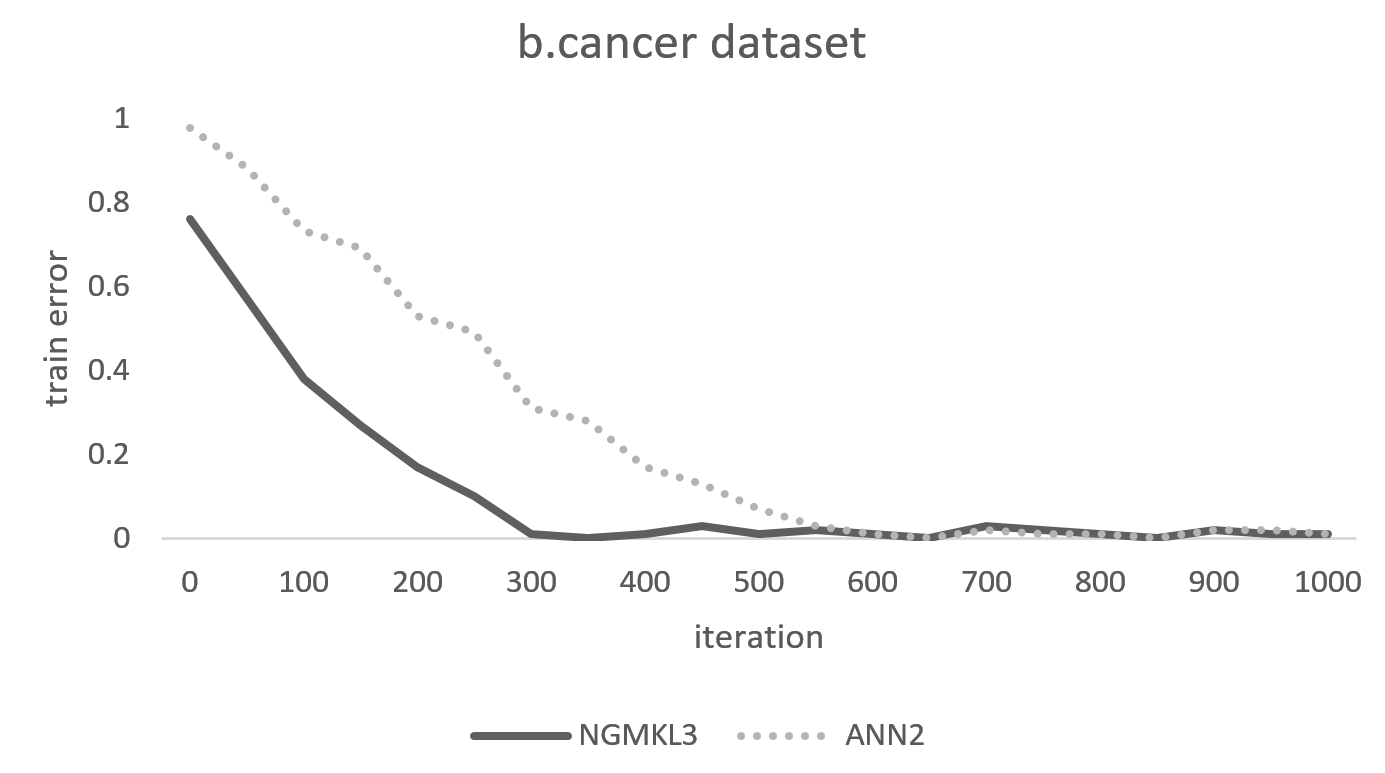}
    \end{minipage}
     \hfill
     \begin{minipage}[b]{0.45\textwidth}
         \centering
         \includegraphics[width=\textwidth]{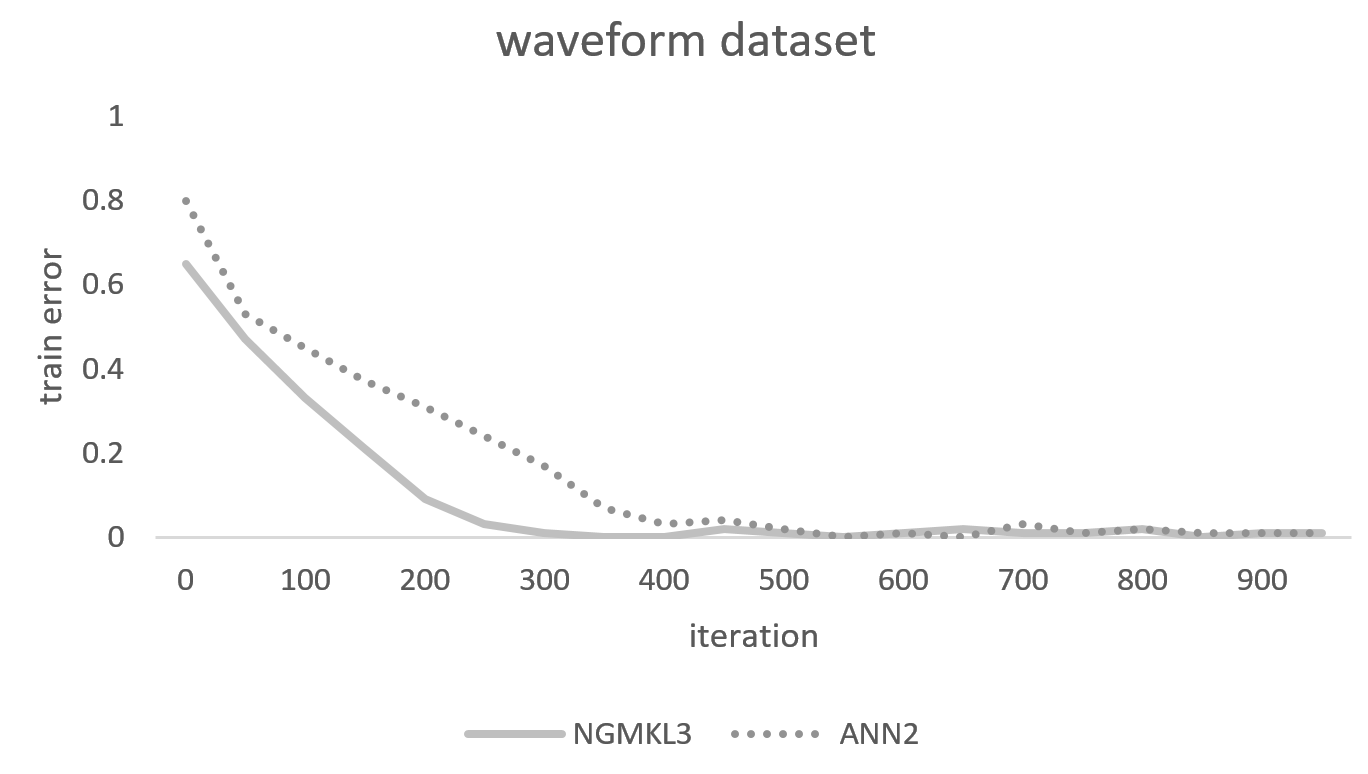}
     \end{minipage}
    \caption{The number of iterations to train NGMKL3 and ANN2 on b.cancer (left subfigure) and waveform (right subfigure) datasets.}
    \label{fig:iterations}
\end{figure*}

In Figure.~\ref{fig:NNMKL}, we compare the performance of NGMKL3 with two Artificial Neural Networks (ANN) with two layers. The architecture of ANN, including the layers, neurons within each layer, and the activation functions, is the same as that of NGMKL3. In order to configure and train the ANN, two approaches can be considered. One approach is to use initial weights from MKL and freeze them during training (ANN1), and the second is to use random initial weights and train the whole neural network (ANN2). The reason behind the first approach is to demonstrate the importance of the kernel layer and provide a fair comparison of MKL methods with neural networks. As seen in Figure.~\ref{fig:NNMKL}, in all datasets, ANN1 and NGMKL3 have almost the same classification error rate, but ANN2 performs worse than both approaches in terms of recognition accuracy.
In the next experiment, we compare the training classification error of NGMKL3 and ANN2. According to our results, a lower number of iterations is required for the convergence of NGMKL3 compared to ANN2. In other words, by transforming data from the kernel layer, a rich representation of data is formed, allowing us to capture underlying patterns with fewer training iterations. In Figure.~\ref{fig:iterations}, we depict the average training error based on the iteration of ANN2 and NGMKL3 for "b.cancer" and "waveform" datasets. In the "b.cancer" dataset, for instance, NGMKL3 achieves a near-zero training error after 300 iterations, while ANN2 reaches this level after 600 iterations.  

\section{Conclusion}
\label{sec:conclusion}
   In this paper, we showed that the framework of the MKL could be interpreted as a neural network with one hidden layer in which the input data is represented by a set of kernels. We proposed NGMKL, a neural generalization of the MKL problem, using this interpretation. Like conventional neural networks, and unlike typical MKL algorithms, NGMKL employs the softmax layer for classification. Furthermore, there are no constraints on the parameters, and all the parameters are trained simultaneously. Moreover, It leverages nonlinear transformations, and the architecture can be extended to multiple layers. We evaluated NGMKL on the UCI datasets and compared it with typical MKL algorithms. The experimental results show that the proposed model can boost the performance of MKL models.

\bibliography{refs}

\end{document}